\documentclass[11pt]{article}
\usepackage{acl2012}
\usepackage{times}
\usepackage{latexsym}
\usepackage{amsmath}
\usepackage{multirow}
\usepackage{url}

\title{Detecting English Writing Styles For Non Native Speakers}

\author{Yanging Chen, Rami Al-Rfou', Yejin Choi \\
  Department of Computer Science \\
  Stony Brook University \\
  NY 11794, USA \\
  {\tt \{cyanqing, ralrfou, ychoi\}@cs.stonybrook.edu}}

\date{12/15/2011}

\usepackage{graphicx}
\begin{document}
\maketitle
\begin{abstract}

This paper presents the first attempt, up to our knowledge, to classify English
writing styles on this scale with the challenge of classifying day to day language written by writers with different backgrounds covering various areas of topics.The paper proposes simple machine learning algorithms and simple to generate features to solve hard problems. Relying on the scale of the data available from large sources of knowledge like Wikipedia. We believe such sources of data are crucial to generate robust solutions for the web with high accuracy and easy to deploy in practice. The paper achieves 74\% accuracy classifying native versus non native speakers writing styles.

Moreover, the paper shows some interesting observations on the similarity between different languages measured by the similarity of their users English writing styles. This technique could be used to show some well known facts about languages as in grouping them into families, which our experiments support.
\end{abstract}

\section{Introduction}
%Rami's Part
The internet nowadays is more diverse than any time before, with the introduction of social networks the majority of users are not any more native English speakers. This puts more challenges on the services providers to accommodate the English content to the new users. This paper tackles the challenge of identifying the native language of the user from their writing styles. We believe this task as a first step will be crucial in the development of many useful applications.

Wikipedia is well known source for knowledge. Recently, it is used extensively to help in solving different information retrieval tasks especially the ones that involves semantic aspects. The use of wikipedia can be expanded to help the common NLP tools to perform better with the help of the diversity of topics and authors of wikipedia pages. Which will help in the data sparsity problem. The sustained growth of the content of wikipedia can bring performance gains with no much additional costs.

The detection of the writer's native language can be helpful in application that targets new learners of English as a second language. Moreover, it could be adapted to transcribed text to help better voice recognition application when dealing with the non native speakers accents.

\section{Related Work}
%Chen's part
%Mentions what people did, and how our work is different and original.

The first work related with native language identification is that of \cite{koppel2005automatically}, in which they tried profiling anonymous authors with their native languages. Totally five different groups of English authors (whose native languages are Russian, Bulgarian, French, and Spanish) were picked from the first version of {\em International Corpus of Learner English} (ICLE) in their experiments. By applying a combined feature sets, including function words, character n-grams, part-of-speech bi-grams and spelling mistakes, they gained an accuracy of 65\% if considered style features only. These results suggested that syntactic features are valuable when trying to categorize authors by their native languages. Also in \cite {koppel2005determining}, they considered not only letter n-grams and funciton words but errors and idiosyncrasies, including orthography errors, syntax errors, neologisms and part-of-speech bigrams errors. Finally the accuracy on classifying authors from five differents countries can reach above 80\%. \cite {argamon2009automatically} concluded some more important features in the task of profiling authors of an anonymous text.

Similar work was done by \cite {tsur2007using}. They focused on the relationships between choice of words in second language writing and the frequency of native language syllables, also known as the phonology of native languages. \cite {estival2007author} studied a wide range of lexical and document structure features in their native languages classification task. And \cite {zheng2003authorship}, though they did not directly conduct related experiments on nationality detection, they provide some features of style markes that could be used in the task of judging one's native languages. Besides, \cite {gamon2004linguistic} analysized the power of some general features under different frequency cutoffs. But none of these measured the usefulness of syntactic features under a general condition for the task of native language detection.

\cite {wong2009contrastive} replicated the work of \cite {koppel2005automatically} and digged more in the field of syntactic structures. They experimented on three selected syntactic errors, which are commonly observed in non-native English Users, including subject-verb disagreement, mismatch of noun-number pairs and wrong usage of determiners and the best overall accuracy was 73.71\% on the second version of ICLE across seven languages. \cite {wong2010parser} first considered applying parser features in the task--though these features are hard to extract compared with other syntactic features. What's more, \cite {wong-dras:2011:EMNLP} continued their works in native language detection and focused more on the influence of syntactic structures, specifically parsing trees. They tried to exploiting the parsing structures by applying Standford parsers and C\&J parsers with different parameters to certain corpus, and capture the number of usages of some distinguishable rules. Their results and observations suggested that the syntactic structures would be supportive in detecting native languages and improving the performance of existing classifiers.

Different from previous works mentioned above, our task runs on a totally different platform--wikipedia. Our goal is to find out the influence of one's native languages on the style of his/her writings under the circumstance of talking and discussion. With the help of huge amount of available data, we can try exploring the statistics features of a certain languages using similar features in \cite {koppel2005automatically} and \cite {wong-dras:2011:EMNLP}, as well as the distribution of part-of-speech(PoS) n-grams and word n-grams.      

\section{Wikipedia}
%Explain why wikipedia is an awesome source
%Explain wikipedia structure
%Explain how there are two different ways to get users contributions
%Explain the comments extraction algorithm
%Rami's Part

Wikipedia is the de facto source of knowledge for internet users. Wikipedia is the 5\textsuperscript{th} most popular website according to Google ranking. For researchers Wikipedia is a giant linguistic and social jar of experiments. The richness of the website content that is written by users from different backgrounds represents a robust sample of the current languages usage by native and non native speakers.

With more than 90 thousand active users and 4.4 million article the content of Wikipedia spans large number of topics. The diversity of the authors of those articles beside the records of the revisions that are stored in a database of revisions that the website offer for free presents a realistic source of text. Such resource presents a higher quality of data that is not achievable by the other commonly used sources of text as news and scientific papers.

\begin{figure}[htp]
\centering
\includegraphics[scale=0.60]{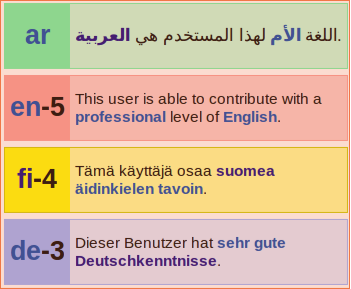}
\caption{Wikipedia languages skills info box (Babel)}
\label{babel}
\end{figure}

Such successful website has a complex database structure to serve its users. Therefore, extracting data could be a complex process. Our goal is to identify the languages skills of the users and collect their contributions. To achieve the first task Wikipedia has a an information box called \emph{Babel} that users can add voluntarily to their profile pages to state their skills in different languages. Figure \ref{babel} shows a user who identified his native language and her skills in 3 other non native languages in a scale for 1-5. This info box will be indexed in the database as categories.

\begin{figure*}[htp]
\centering
\includegraphics[scale=0.285]{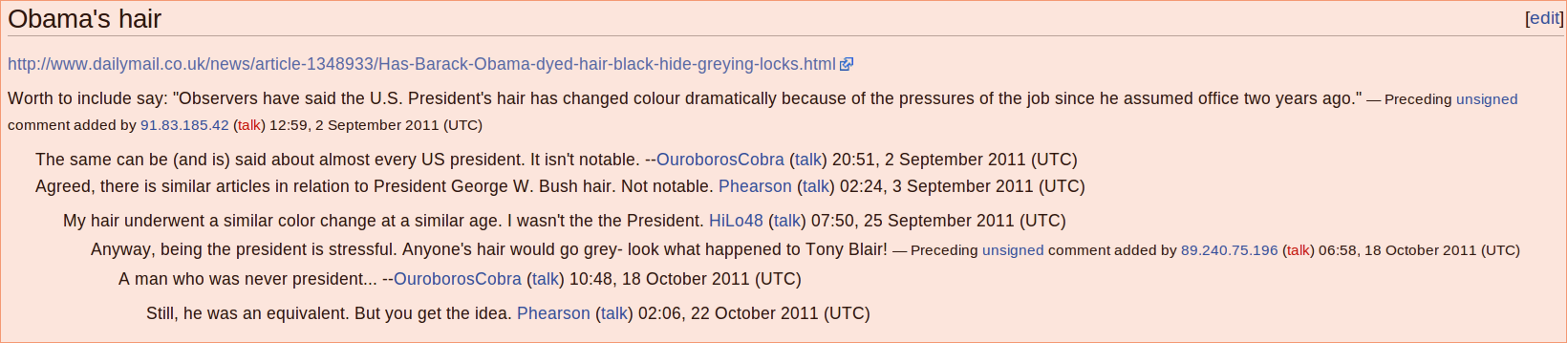}
\caption{Example of a conversation in the discussion pages}
\label{obama}
\end{figure*}

To collect the contributions of a specific user the task is more complex procedure. The diffs between Wikipedia pages revisions has to be generated and linked back to the user table. However, the resources we have to process such huge amount of data did not allow us to do that\footnote{Recent efforts were made to generate the diffs \url{http://dumps.wikimedia.org/other/diffdb/}}. Instead we noticed that Wikipedia pages have accompanying discussion pages where users discuss different aspects of the articles. In those pages the tradition is to sign the user comments with a signature that link back to the user. Figure \ref{obama} shows the style of the writing of those talk pages are less formal and technical than the main pages of Wikipedia and has more conversational stylistic features.

The patterns of the recommended signatures styles are limited in number, however, in practice the users use various patterns that makes the detection rules ambiguous. The detection algorithm implemented relies on complex regular expressions and applies best effort strategy.

\section{Experiments}
\begin{figure}[htp]
\centering
\includegraphics[scale=0.5]{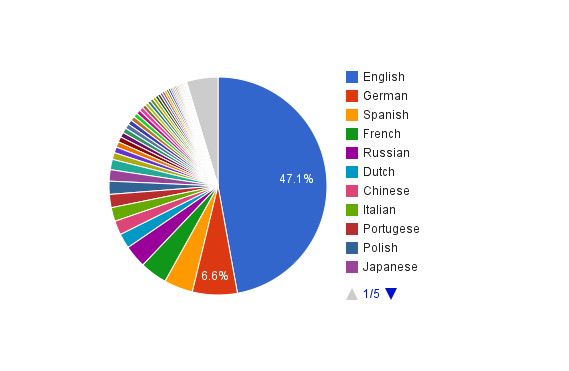}
\caption{Users distribution over native languages in English Wikipedia}
\label{native_dist}
\end{figure}

%Rami's part
We found that around 60 thousands users specified their language skills. Figure \ref{native_dist} shows that the percentage of users who claimed that their one of their native languages is English is around 47\% of English wikipedia users base. 

We parsed the talk pages with the \verb+namespace=1+, they represents x\% of the talk pages, which produced around 12 million comments. Only 2.4 million comment we could identify to users with known language skills. Moreover, not all the users made comments in the talk pages we parsed, therefore, The number of the users who made at least one contribution in the extracted contributions is around 30 thousand user.

As we have large number of comments and users and as we believe the data we have is still noisy. We applied the following filtering mechanisms:
\begin{itemize}
\item We picked the users of the most popular 19 native languages.
\item We picked out of the English native speakers the users who specified the EN-US as their native language only to avoid users who are so skilled in English but are not living in English speaking countries.
\item We excluded the users who specified more than one native language out of the picked native languages to avoid unrealistic scenarios where users claim to be native in more than two languages.
\end{itemize}

The new data set after the filtering is consistent of 9857 user and 589228 comments.

\subsection{Setup}
%Explain the pruning and the filtering that was done to the data.
The following experiments are conducted under the following conditions:
\begin{itemize}
\item The accepted comments has to have at least 20 tokens to avoid short and non meaningful comments.
\item Proper nouns are replaced by their tags to avoid bias toward topics.
\item Non ascii characters are replaced by a special character to avoid bias foreign languages usage in the comments.
\item The classifier has balanced number of comments for each its classes. Therefore, the two baseline classifiers; the most common label and the random classifier will have an accuracy of 1/number of classes.
\item Logistic Regression algorithm is used to for the classification task.
\item The data set is split to 70\% training set, 10\% development set and 20\% testing set.
\end{itemize}

\subsection{Features}
The comments of training set is grouped by class and the following frequency distribution are calculated for each class:
\begin{itemize}
\item 1-4 grams over the comments words.
\item 1-4 grams over the characters of the words of the comments.
\item 1-4 grams over the part of speech tags.
\end{itemize}

For each comment ($C$) similarity measurements($Sim$) are calculated against each n-gram frequency distribution ($f(n)$) according to the following equations:

\[
  count(x, f, n) = \left\{ 
  \begin{array}{l l}
    FreqDistCount(x, f, n), & \\\quad \text{if $x$ is in $f(n)$}\\
    1,&\\\quad \text{if $x$ is not seen before}\\
  \end{array} \right.
\]
\[
  Sim(C,f,n) = \sum_{x \in ngrams(C,n)} \log_2 (count(x,f,n))
\]

So if our problem has six classes this will generate $6*3*4 = 72$ features.

Other features also included the relative frequency of each of the stop words to the size of the comments. The  125 stop words are extracted from the NLTK stop words corpus. Moreover, the average size of words and the average number  of sentences is also added.

\subsection{Popular Languages Experiment}
\begin{figure}[htp]
\centering
\includegraphics[scale=0.45]{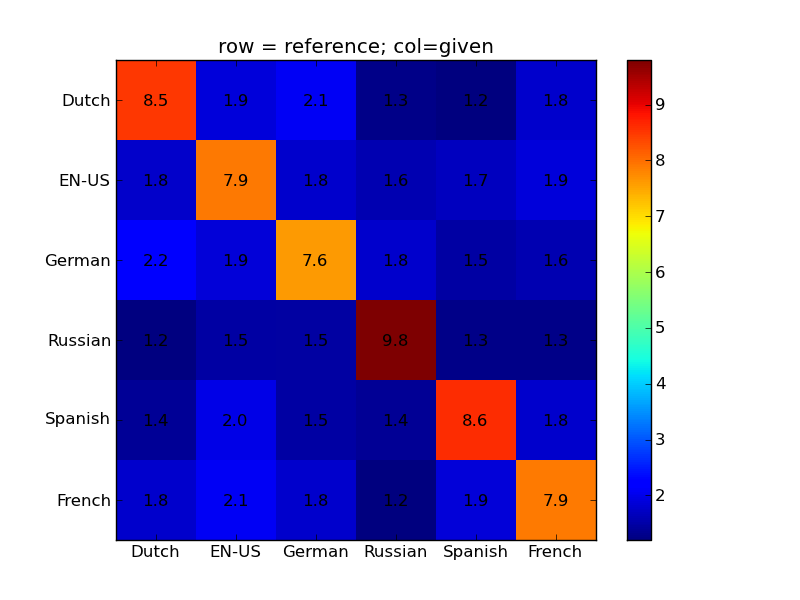}
\caption{Popular Languages experiment confusion matrix}
\label{pop_cfm}
\end{figure}

%Explain the experiment and the results
The most popular six languages: US-EN, German, Spanish, French, Russian and Dutch  are chosen to train a classifier to detect the user's native language. Figure \ref{pop_cfm} shows the confusion matrix of the experiment that is done using 100\% of data set, about 150K comment. We can see clearly that the Russian users are the easiest to identify. Moreover, the classifier is the most confused distinguishing the German and the Dutch users with error $>2.0\%$ and to a less degree between (EN-US, French) and (EN-US, Spanish). These numbers confirm a basic intuition that users who have geographical proximity will have more borrowed words and grammars between their native languages which will affect their writing styles in English.

\begin{figure}[htp]
\centering
\includegraphics[scale=0.45]{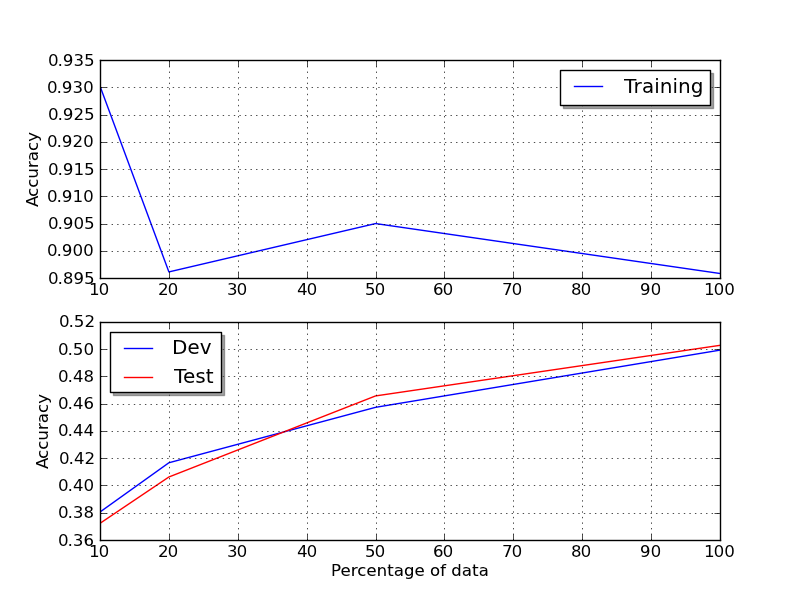}
\caption{Popular languages experiment learning curves}
\label{pop_lc}
\end{figure}

Figure \ref{pop_lc} shows that the best accuracy that the classifier achieved is 50.275\%. The learning curves shows a typical over fitting situation where the more data you have the better the classifier can achieve. And here the size of data that can be extracted from wikipedia plays a significant role to boosts the accuracy from 37\% to over 50\%. The growth of the curve is similar to $\sqrt{x}$ curve which suggests the importance of the increase in the coverage of unique words that the frequency distribution which grows also with the same rate.

\subsection{Languages Families Experiment}
%Explain the experiment and the results

The confusion in classifying Dutch and German users suggests that there is a similarity between groups of languages. Referring to the linguistics research history of classifying the languages into families according to similar features and development history, this experiment tries to put such grouping under the microscope. 17 languages are grouped into 5 families as the following:
\begin{itemize}
\item North Germanic\\
German, Dutch, Norwegian, Swedish, Danish
\item Roman\\
Spanish, French, Portuguese, Italian
\item Uralic \\
Russian, Polish, Finnish, Hungarian
\item Asian\\
Chinese, Japanese, Korean
\item English
\end{itemize}

\begin{figure}[htp]
\centering
\includegraphics[scale=0.45]{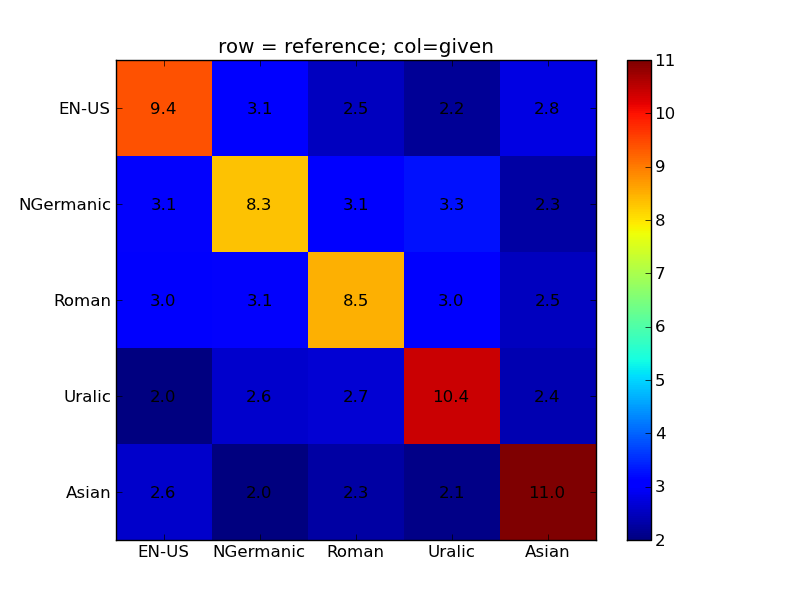}
\caption{Languages families experiment confusion matrix}
\label{fam_cfm}
\end{figure}

Figure \ref{fam_cfm} shows the that the Asian native speakers has a clear style of writing English that is easy to detect relatively. Moreover, as English belongs to the Anglic language family which also belongs to the West Germanic family of languages, we can see that clearly in terms of the high error rates $>3.0\%$. Other trends can be explained according to the users geographical proximity as in cases of (Uralic, North Germanic), (North Germanic, Roman).

The learning curves of this experiment are similar to the ones from the previous experiment with the exception that the best accuracy that is achieved is less with $47.542\%$ when 100\% of the data is used. That could be explained when we notice that the total number of comments that are used in this experiment is 82K which around 50\% of the data used in the previous experiment. With 50\% of the data in the popular languages experiment the classifier achieved similar performance. We could not add more comments because of the constraints that all the classes should have the same number of comments.

\begin{figure}[htp]
\centering
\includegraphics[scale=0.45]{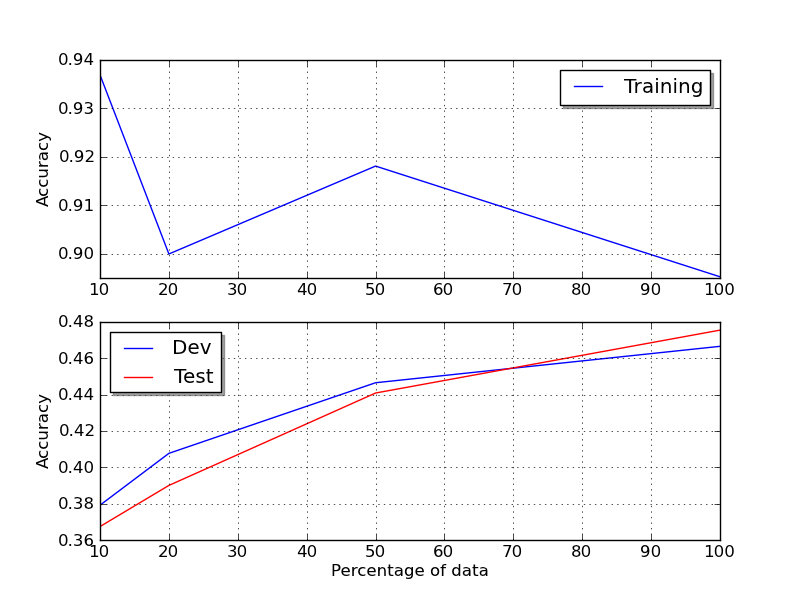}
\caption{Languages families experiment learning curves}
\label{fam_lc}
\end{figure}

\subsection{Native vs Non Native Experiment}
\begin{figure}[htp]
\centering
\includegraphics[scale=0.45]{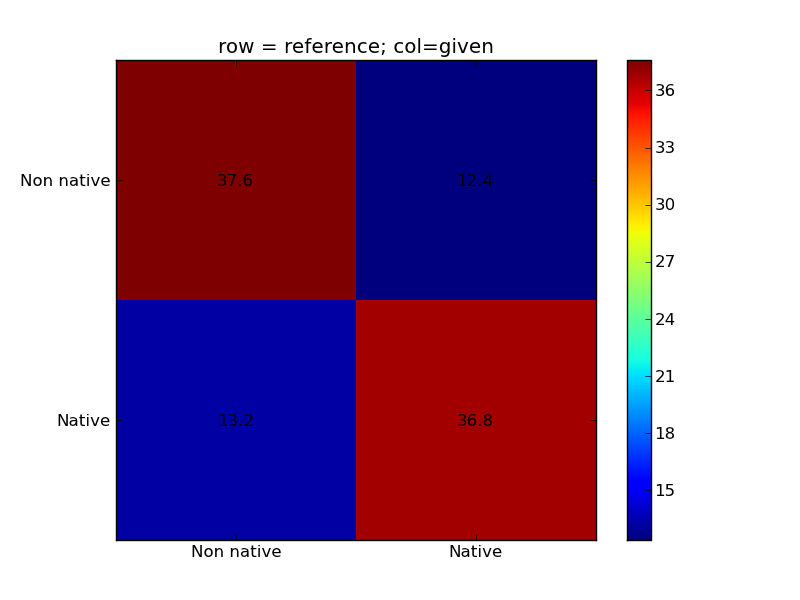}
\caption{Native vs Non Native speakers experiment confusion matrix}
\label{non_cfm}
\end{figure}

\begin{figure}[htp]
\centering
\includegraphics[scale=0.45]{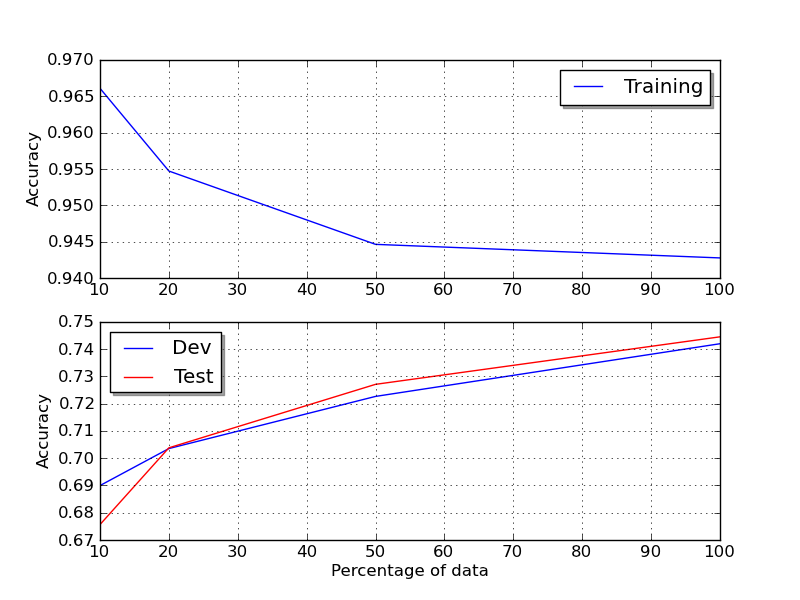}
\caption{Non native vs native experiment learning curves}
\label{non_lc}
\end{figure}

%Explain the experiment and the results
In this experiment all the non native English speakers were labelled as Non native. Figure \ref{non_cfm} and figure \ref{non_lc} shows that the classifier could achieve $74.449\%$ accuracy using around 322K comment divided between training, development and testing sets. Such high accuracy makes the classifier able to be deployed for practical usages.

\section{Writing Styles}
%Chen's part
Another type of Experiment focuses on the usage of PoS n-grams, trying to distinguish users by comparing the similarity of the PoS-ngrams distribution with a candidate language. We use the same definition of "similarity" as mentioned in the previous section and \cite {alpaydin2004introduction} and build a 20-classifier based on the whole training data (524 MB). After training our language model, The accuracy was around 95\% on training data.

The two baselines we selected are: Baseline-Max and Baseline-Random. Where in Baseline-Max we simply put each user into the category of native US English speaker since this class has the largest size with the
probability of about 34\%. And in Baseline-Random, we choose randomly from 20 candidate languages. The expectation of accuracy should be 5\%.

Since there exist some PoS-ngrams that might never appear in the training data, which means we have not seen such PoS-ngram before in any of the candidate category. We simply define these comments as "zero comments"--it is not precise to measure the value of these unseen PoS n-grams and we should put it into the category of native US English speaker to maximize the probability of correctness. Also, it sounds not reasonable if we consider comments that are too short considering the experiments mentioned in \cite {wagner2009judging}. So we first try on comments that has more than 100 PoS n-grams, which means the comment has more than (100+n-1) tokens.

\begin{table}[h]
\begin{center}
\small\addtolength{\tabcolsep}{-5pt}
\begin{tabular}{|l|l|l|l|l|}
\hline & \bf Overall & \bf Nonzero & \bf Available & \bf Nonzero \\ 
& \bf Accuracy & \bf Accuracy & \bf Count & \bf Count \\ \hline
4-grams & 35.47\% & 37.54\%  & 13067 & 8196 \\
tri-grams & 25.99\% & 25.96\% & 13209 & 13162 \\
bi-grams & 15.77\% & 15.77\%  & 13375 & 13375\\
uni-grams & 9.93\% & 9.93\%  & 13569 & 13569 \\
baseline-max & 34.03\% & 34.01\% & 13067 & 13067 \\
baseline-random & 5.00\% & 4.96\% & 13067 & 13067 \\
bag-of-words & 33.98\% & N/A & 13569 & 0 \\
\hline
\end{tabular}
\end{center}
\caption{\label{font-table} Result of Different PoS n-grams. }
\end{table}

It is clear from the table that the higher level of PoS n-grams, the higher the accuracy. But The experiment on 4-grams shows that about 1/3 of these comments contains unseen PoS 4-grams. Our training data contains more than 300,000 possible 4-grams in the category of native US English speaker, but that is still not enough, not to mention that the category of Korean native speaker only covers 18,000 4-grams. It seems that a good estimation of unseen PoS 4-grams can boost the accuracy. What's more, even word-unigram suffers the same problem (each comment with length greater than 100 had a token that was not seen anywhere before).

Another experiment runs on different length of comments using 4-grams, and we believe that the longer the comments, the higher the accuracy.

\begin{table}[h]
\begin{center}
\small\addtolength{\tabcolsep}{-5pt}
\begin{tabular}{|l|l|l|l|l|}
\hline Accuracy & \bf len$>$50 & \bf len$>$100 & \bf len$>$150 & \bf len$>$200 \\ \hline
4-grams & 33.18\% & 35.47\%  & 36.61\% & 37.57\% \\
baseline-max & 34.03\% & 34.01\% & 34.36\% & 35.04\% \\
baseline-random & 5.00\% & 4.96\%  & 5.01\% & 4.27\%\\
\hline
\end{tabular}
\end{center}
\caption{\label{font-table} Result of varies length overall. }
\end{table}

\begin{table}[h]
\begin{center}
\small\addtolength{\tabcolsep}{-5pt}
\begin{tabular}{|l|l|l|l|l|}
\hline Accuracy & \bf len$>$50 & \bf len$>$100 & \bf len$>$150 & \bf len$>$200 \\ \hline
4-grams & 33.40\% & 37.54\%  & 40.07\% & 42.24\% \\
baseline-max  & 34.75\% & 35.26\% & 36.15\% & 37.85\% \\
baseline-random  & 5.04\% & 4.67\% & 4.97\% & 4.22\% \\
\hline
\end{tabular}
\end{center}
\caption{\label{font-table} Result of varies length on nonzero data. }
\end{table}

We also observed that, some frequently appeared 4-grams occupy rather different portion in the distribution. For instance,
(IN,DT,NN,PRP): 0.13\% in Portugal but only 0.04\% in Korean
(NN,NN,IN,DT): 0.15\% in Portugal but only 0.05\% in Polish
(TO,DT,NN,IN): 0.11\% in Arabic while less than 0.06\% in any other languages
(',',CD,NNP,CD) and (NNP,CD,-LRB-,NNP): Appeared 10 times more in Korean than other languages, especially Hungarian.
(NN,PRP,VBZ,RB): Japanese and danish users prefer to use this.
 
Penalty for UNSEEN PoS-ngrams is too high and most of the time the correct answer was filtered at the first round, which means we never had a chance to calculate the similarity of distribution for the candidate. We also tried to apply different cutoffs in the experiment, which means we only count some frequent PoS-grams (without considering rare PoS n-grams) and avoid the possible spikes of weights in the process of learning. We tried to focus on most frequent k PoS 4-grams (k=100, 500, 2000, 5000) and those 4-grams appeared in more than k different candidate languages (k=10,15,20). As mentioned before, UNSEEN PoS-ngrams has some power in deciding some candidates languages, but the thresholds we found seems not as good as we expected.

\begin{table}[h]
\begin{center}
\small\addtolength{\tabcolsep}{-5pt}
\begin{tabular}{|l|l|}
\hline Accuracy & \bf 4-grams \\ \hline
no cutoff  &  37.54\% \\
most frequent 100 & 12.81\% \\
most frequent 500 & 15.26\% \\
most frequent 2000 & 21.76\% \\
most frequent 5000 & 30.55\% \\
appeared in 10 lang & 28.93\% \\
appeared in 15 lang & 32.27\% \\
appeared in all lang & 30.14\% \\
\hline
\end{tabular}
\end{center}
\caption{\label{font-table} Result of different cutoffs on PoS-ngrams }
\end{table}

To measure how ofter we throw away a correct answer, we eliminated all comments that contains unseen n-gram in the category of correct answer. After the tricky operation, we discarded about 3/4 of the available testing data, but we found a different result in the accuracy. In our tricky data, only 545 out of 2265 comments has unique candidate, and we got an accuracy of 78.41\% if the correct answer is not eliminated due to unseen PoS n-grams and probably compare between the candidate native language with US English. This phenomenon shows that the language model is reliable as long as we can calculate the distribution similarity. What's more, our model has a property of high precision and low recall. In our tricky data, the distribution of models output is:  

\begin{table}[h]
\begin{center}
\small\addtolength{\tabcolsep}{-5pt}
\begin{tabular}{|l|l|l|l|}
\hline & \bf Actual & \bf Predicted & \bf Correct \\ 
& \bf Occurrences & \bf Appearances & \bf Prediction \\ \hline
Deutsch & 181 & 150 & 78 \\
Japanese & 2 & 2 & 2 \\
Polish & 10 & 19 & 9 \\
Mandarin & 14 & 18 & 12 \\
Turkish & 4 & 4 & 4 \\
Finnish & 1 & 3 & 1 \\
Cantonese & 1 & 1 & 1 \\
Arabic & 14 & 13 & 13 \\
Danish & 11 & 18 & 10 \\
Hungarian & 4 & 4 & 4 \\
Spanish & 86 & 90 & 39 \\
Portuguese & 172 & 169 & 169 \\
French & 32 & 44 & 12 \\
Netherlands & 103 & 203 & 73 \\
US English & 1485 & 1329 & 1232 \\
Korean & 0 & 0 & 0 \\
Italian & 11 & 14 & 11\\
Swedish & 36 & 50 & 30 \\
Norwegian & 12 & 25 & 12 \\
Russian & 86 & 109 & 64 \\
\hline
\end{tabular}
\end{center}
\caption{\label{font-table} Statistics on tricky data }
\end{table}

while in the real data, we get hundreds of non-native speakers falling into the category of US English native speaker since this class covers more n-grams than the others.

In order to deal with unseen ngrams, we apply another method that can estimate the occurrance of a certain 4-gram by cascade down to using Trigram/Bigram/Unigram. But these attempts seems not Performing well, with only trigram/bigram estimation, the accuracy drop down to 34.5\%, and using only bigram/unigram estimation provide an accuracy of 29.7\%. If we apply both two strategies, the accuracy is 31.3\%. For most of the cases, cascade helps reduce the number of we predict some one as US English native, but it seldom solve the problem if the appearance of n-grams is really small, for instance, in Korean and Japanese.

\section{Conclusions}
%Chen's Part
Our experiments show that syntactic structures and writing styles appear to be different for people from different area. Even consider only these features, we can make judgment on one's native languages. And it will be supportive if applying them together with other semantic features.      

\section{Future Work}
%Chen+Rami
As our results shows promising applications and trends using Wikipedia data to solve hard problems in robust means, we are looking to investigate the effects of adding the wikipedia diffs, especially the non minor ones, as another source of user contributions. Moreover, the minimum size of the comments affects the performance of our classifiers, the relation between the quality of the data used and the accuracy of the classification is another interesting aspect.

The languages families experiments suggest the usefulness of using the English writing styles to define the similarities between different languages. This could lead to an interesting explanations and/or observations regarding the origins of some languages as Korean language which stays till a mysterious topic.

Another direction is to solve the over fitting problem in our learning algorithms by applying smarter feature selection and adding more distinguishing features.

Also, we can try more scoring scheme other than pure similarity, for instance, total sum of rankings on all possible n-grams. This method could also be helpful in avoiding the spikes generated by rare n-grams in the training data.

\section*{Acknowledgements}
%Chen's Part
%We should thank Prof. Skiena for the resources he gave us.
We would like to thank Steven Skiena for the discussion and the advice. This work will not be available without the computing resources offered by his lab. We are also indebted to the NLTK and the Sklearn teams for producing excellent NLP and machine learning resources.

\bibliography{myrefs}
\bibliographystyle{acl}

\end{document}